\documentclass[10pt, a4paper]{article}

\usepackage{lrec2026}
\usepackage{graphicx}
\usepackage{booktabs}
\usepackage{multirow}
\usepackage{amsmath}
\usepackage{amssymb}
\usepackage{gensymb}
\usepackage{xcolor}
\usepackage{soul}
\usepackage{enumitem}
\usepackage{tikz}
\usetikzlibrary{arrows.meta, shapes, positioning, fit, backgrounds}
\setlist[itemize]{leftmargin=1.5em, itemsep=4pt, topsep=4pt, parsep=2pt}

\title{ZeroR@CHiPSAL 2026: Two-Stage Vision-Language Adaptation with Contrastive Learning for Nepali Meme Classification}

\name{Nitiz Khanal}

\address{Pulchowk Campus, Institute of Engineering, Tribhuvan University \\
  Lalitpur, Nepal \\
  \texttt{khanalnitij20@gmail.com}}

\abstract{
This paper presents our system for the CHiPSAL 2026 shared task on multimodal hate speech and sentiment detection in Nepali memes. We address both subtasks: binary hate speech classification and three-class sentiment analysis. Our approach adapts the Robust Adaptation of Hateful Meme Detection (RA-HMD) framework using Qwen3-VL-8B-Instruct, a state-of-the-art vision-language model with native Devanagari support. We employ a two-stage training pipeline: (1) LoRA fine-tuning with an MLP projection head for generative classification, and (2) contrastive backbone fine-tuning with supervised InfoNCE loss. We handle class imbalance through minority oversampling, image augmentation, and focal loss. At inference, we ensemble Stage 1 token probabilities with Stage 2 classifier scores using validation-tuned weights. Our end-to-end approach eliminates error propagation from separate OCR and translation pipelines by leveraging the model's native Devanagari understanding. Our system achieved \textbf{2nd place} on hate speech detection (F1: 0.797) and \textbf{4th place} on sentiment analysis (F1: 0.518). We provide detailed ablations, error analysis, and insights into adapting large vision-language models for low-resource South Asian languages.
\\ \newline }

\begin{document}

\maketitleabstract

\section{Introduction}

The proliferation of memes on social media platforms has created new challenges for content moderation systems. Unlike traditional text or image classification, memes require understanding the complex interplay between visual and textual elements, where meaning often emerges from their combination rather than either modality alone \citep{kiela2020hateful}. A seemingly innocuous image paired with specific text can convey hate, while aggressive text on a humorous image might be entirely benign. This multimodal reasoning challenge is further compounded for low-resource languages, where annotated datasets are scarce and pre-trained models have limited exposure to the language and its cultural context \citep{parihar2021hate}.

Nepali, spoken by approximately 32 million people, presents unique challenges for automated content moderation. The language uses Devanagari script, which is embedded directly into meme images rather than appearing as separate text. Cultural references, humor styles, and implicit meanings rooted in Nepali society add layers of complexity that general-purpose models may not capture. Furthermore, the code-mixing of Nepali with English, common in social media, creates additional processing challenges.

The Second Workshop on Challenges in Processing South Asian Languages (CHiPSAL 2026), co-located with LREC-COLING 2026, addresses these challenges through shared tasks focused on South Asian language processing \citep{sarves2026chipsal}. We participated in the multimodal hate and sentiment detection task \citep{thapa2026sharedtask}, which provides carefully curated Nepali meme datasets collected from social media platforms \citep{thapa2025multimodal, thapa2025cross, bhandari2023crisishatemm}. The shared task comprises two subtasks:

\begin{itemize}
    \item \textbf{Subtask A -- Hate Speech Detection}: Binary classification of memes as hateful (content targeting groups based on ethnicity, religion, gender, or caste, or promoting violence/discrimination) or safe.
    \item \textbf{Subtask B -- Sentiment Analysis}: Three-class classification of memes into negative, neutral, or positive sentiment categories.
\end{itemize}

Our approach leverages recent advances in vision-language models (VLMs), specifically adapting the Robust Adaptation of Large Multimodal Models (RA-HMD) framework \citep{mei2025rahmd} for low-resource Nepali content. Rather than constructing complex pipelines involving OCR extraction, machine translation, and separate text/image encoders, we leverage Qwen3-VL-8B-Instruct \citep{qwen2vl}, a state-of-the-art VLM with native support for multiple scripts including Devanagari. This end-to-end approach eliminates error propagation that would occur in pipeline-based systems.

Our key contributions are:
\begin{enumerate}
    \item A two-stage training framework combining generative fine-tuning with contrastive learning, adapted from RA-HMD for low-resource multimodal classification.
    \item Task-specific configurations addressing the distinct challenges of binary hate speech detection versus three-class sentiment analysis.
\end{enumerate}

Our system achieved \textbf{2nd place} on hate speech detection (F1: 0.797) and \textbf{4th place} on sentiment analysis (F1: 0.518), demonstrating the effectiveness of adapting modern VLMs for low-resource multimodal tasks.

\section{Related Work}

\subsection{Multimodal Hate Speech Detection}

The challenge of detecting hate speech in multimodal content gained significant attention with the Facebook Hateful Memes Challenge \citep{kiela2020hateful}, which demonstrated that simple fusion of unimodal representations is insufficient. The challenge showed that even state-of-the-art vision and language models struggled when hate was conveyed through subtle interactions between image and text.

Subsequent work has explored various fusion strategies, from early fusion approaches that concatenate features to more sophisticated cross-modal attention mechanisms. \citet{pramanick2021momenta} introduced methods for detecting harmful memes and identifying their targets, while \citet{bhandari2023crisishatemm} extended multimodal hate speech analysis to crisis contexts with the CrisisHateMM dataset, providing annotation schemas for directed and undirected hate.

More recently, \citet{mei2024retrieval} proposed retrieval-guided contrastive learning for hateful meme detection, using similar training samples to guide representation learning. This work was extended by \citet{mei2025rahmd} with the RA-HMD framework, which combines LoRA fine-tuning with rationale-aware contrastive learning, achieving state-of-the-art results on multiple English hateful meme datasets including HatefulMemes, MAMI, and PrideMM. Our work adapts this framework for low-resource Nepali content.

\subsection{Vision-Language Models}

The emergence of large vision-language models has transformed multimodal understanding. Early approaches like CLIP \citep{radford2021clip} learned joint image-text representations through contrastive learning on web-scale data. More recent instruction-tuned models like LLaVA \citep{liu2024visual} and the Qwen-VL family \citep{qwen2vl} can follow complex multimodal instructions and generate detailed responses.

Qwen3-VL-8B-Instruct, which we use as our backbone, represents a significant advancement in multimodal understanding. The model supports dynamic image resolution, multiple languages including those using non-Latin scripts, and demonstrates strong performance across diverse vision-language benchmarks. Critically for our application, it has native support for Devanagari script, enabling end-to-end processing of Nepali text embedded in images without requiring separate OCR.

\subsection{Parameter-Efficient Fine-Tuning}

Fine-tuning large models on downstream tasks is computationally expensive and can lead to catastrophic forgetting. Low-Rank Adaptation (LoRA) \citep{hu2022lora} addresses this by freezing pre-trained weights and injecting trainable rank decomposition matrices into transformer layers. This approach reduces trainable parameters by orders of magnitude while maintaining competitive performance.

For vision-language models, LoRA has proven particularly effective, allowing adaptation to new tasks and domains without the computational cost of full fine-tuning. We apply LoRA to all linear projections in Qwen3-VL, enabling efficient adaptation to Nepali meme classification.

\subsection{Contrastive Learning for Classification}

Supervised contrastive learning \citep{khosla2020supervised} extends the self-supervised contrastive paradigm by leveraging label information. The objective pulls together representations of samples from the same class while pushing apart representations from different classes. The InfoNCE loss \citep{oord2018representation} provides a principled framework for this objective.

For content moderation tasks, contrastive learning helps learn discriminative representations that capture subtle differences between classes. \citet{mei2024retrieval} and \citet{mei2025rahmd} demonstrated its effectiveness for hateful meme detection, and we adapt this approach for Nepali content.

\subsection{Low-Resource Language Processing}

Processing low-resource languages presents unique challenges due to limited training data, fewer pre-trained resources, and distinct linguistic characteristics. Large language models have shown promise for such languages through cross-lingual transfer \citep{thapa2025large}, but multimodal tasks in low-resource settings remain underexplored.

The CHiPSAL workshop series \citep{sarves2026chipsal} aims to address these gaps for South Asian languages. Prior work on Nepali NLP has focused primarily on text classification and named entity recognition, with multimodal analysis receiving limited attention. Our work contributes to this emerging area by demonstrating effective adaptation of state-of-the-art VLMs for Nepali multimodal content.

\section{Dataset and Task Description}

\subsection{Dataset Overview}

The shared task provides Nepali meme datasets collected from various social media platforms \citep{thapa2025multimodal, thapa2025cross}. The memes contain Devanagari text embedded in images, often with code-mixing between Nepali and English. The annotation follows established protocols for multimodal hate speech \citep{bhandari2023crisishatemm}.

\subsection{Subtask A: Hate Speech Detection}

The hate speech dataset contains approximately 1,068 memes with binary labels:
\begin{itemize}
    \item \textbf{Hate (1)}: Content that attacks, demeans, or incites hatred against individuals or groups based on protected characteristics including ethnicity, religion, gender, caste, nationality, or disability. This also includes content promoting violence or discrimination.
    \item \textbf{Safe (0)}: Content that does not contain hate speech, including neutral information, positive messages, and humor that does not target protected groups.
\end{itemize}

The dataset exhibits class imbalance, with safe memes being more frequent than hateful ones. This imbalance reflects the natural distribution of content on social media but poses challenges for classification.

\subsection{Subtask B: Sentiment Analysis}

The sentiment dataset contains memes labeled with three categories:
\begin{itemize}
    \item \textbf{Negative (0)}: Content expressing criticism, mockery, sadness, anger, frustration, disappointment, or pessimism.
    \item \textbf{Neutral (1)}: Factual information, observations, or content without clear emotional valence.
    \item \textbf{Positive (2)}: Content expressing joy, humor, encouragement, celebration, love, hope, or optimism.
\end{itemize}

The neutral class dominates the distribution, creating a significant class imbalance challenge. Additionally, the boundary between categories can be subjective, particularly for humorous content that may express negativity in a playful manner.

\subsection{Evaluation Metric}

Both subtasks use macro F1-score as the primary evaluation metric:
\begin{equation}
\text{Macro F1} = \frac{1}{C} \sum_{c=1}^{C} \text{F1}_c
\end{equation}
where $C$ is the number of classes and $\text{F1}_c$ is the F1-score for class $c$. This metric gives equal weight to each class regardless of frequency, making it particularly appropriate for imbalanced datasets.

\subsection{Data Preparation}

We create an 85/15 stratified train/validation split (random seed 42) to tune hyperparameters and ensemble weights. Stratification ensures that class proportions are preserved in both splits. All images are converted to RGB format for consistency. We do not use any external data or pre-training beyond the provided training set and the base Qwen3-VL model.

\section{Methodology}

\begin{figure}[htbp]
\centering
\begin{tikzpicture}[
    box/.style={rectangle, rounded corners=2pt, text centered, font=\scriptsize},
    inputbox/.style={box, draw=black!70, fill=gray!10,
                     minimum width=3.2cm, minimum height=0.45cm},
    stagebox/.style={box, minimum width=1.5cm, minimum height=0.4cm,
                     text centered, font=\scriptsize},
    ensemblebox/.style={box, draw=black!80, fill=orange!10,
                        minimum width=3.4cm, minimum height=0.45cm},
    arrow/.style={-{Stealth[scale=0.6]}, semithick, draw=black!60},
    lab/.style={font=\tiny\bfseries}
]

\node[inputbox] (input) {\textbf{Input:} Meme + text};
\node[inputbox, below=0.22cm of input]  (vlm)    {\textbf{Qwen3-VL + LoRA}};
\node[inputbox, below=0.22cm of vlm]    (hidden) {$\mathbf{h} \in \mathbb{R}^{4096}$};

\node[stagebox, draw=teal!70!black, fill=teal!8,
      below=0.55cm of hidden, xshift=-1.1cm] (lmhead) {LM Head};
\node[lab, teal!70!black, above=0.05cm of lmhead, xshift=-0.4cm] {Stage 1};
\node[stagebox, draw=teal!70!black, fill=teal!12,
      below=0.22cm of lmhead] (s1out) {$p^{(1)}$};

\node[stagebox, draw=violet!70!black, fill=violet!8,
      below=0.55cm of hidden, xshift=+1.1cm] (mlp) {MLP$\,{\to}\,\mathbf{z}$};
\node[lab, violet!70!black, above=0.05cm of mlp, xshift=+0.4cm] {Stage 2};
\node[lab, black!60, right=0.05cm of mlp] {$\mathcal{L}_{\mathrm{con}}$};

\node[stagebox, draw=violet!70!black, fill=violet!12,
      below=0.22cm of mlp] (s2out) {Clf $\to p^{(2)}$};

\node[ensemblebox, below=0.55cm of s1out, xshift=+1.1cm] (ensemble)
      {$w \cdot p^{(1)} + (1{-}w)\cdot p^{(2)}$};
\node[lab, orange!70!black, left=0.08cm of ensemble] {\textbf{Ensemble}};

\node[box, draw=black!70, fill=green!10, minimum width=1.8cm,
      minimum height=0.4cm, below=0.25cm of ensemble] (output)
      {$\hat{y} = \arg\max$};
\node[lab, green!50!black, left=0.08cm of output] {\textbf{Output}};

\draw[arrow] (input)  -- (vlm);
\draw[arrow] (vlm)    -- (hidden);

\coordinate (jct) at ([yshift=-0.25cm]hidden.south);
\draw[arrow] (hidden.south) -- (jct);
\draw[arrow] (jct) -| (lmhead.north);
\draw[arrow] (jct) -| (mlp.north);

\draw[arrow] (lmhead) -- (s1out);
\draw[arrow] (mlp)    -- (s2out);

\draw[arrow] (s1out.south) |- ([yshift=-0.28cm]s1out.south)
             -| ([xshift=-0.8cm]ensemble.north);
\draw[arrow] (s2out.south) |- ([yshift=-0.28cm]s2out.south)
             -| ([xshift=+0.8cm]ensemble.north);

\draw[arrow] (ensemble) -- (output);

\end{tikzpicture}
\caption{Architecture overview. Stage~1 uses the LM head for token
probabilities. Stage~2 adds an MLP projection with contrastive loss
($\mathcal{L}_{\mathrm{con}}$) and a classifier. Outputs are ensembled
and the final prediction is made via argmax.}
\label{fig:architecture}
\end{figure}
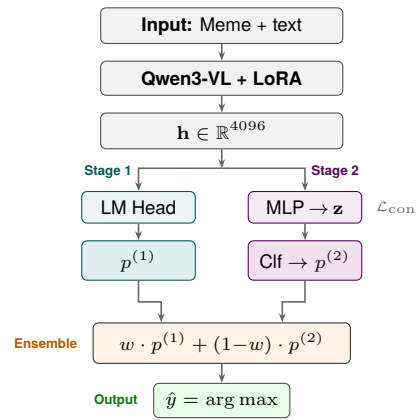

Our system follows a two-stage training pipeline adapted from the RA-HMD framework \citep{mei2025rahmd}. Figure~\ref{fig:architecture} illustrates the overall architecture.

\subsection{Base Model Selection}

We select Qwen3-VL-8B-Instruct as our backbone for several reasons:
\begin{itemize}
    \item \textbf{Native Devanagari support}: The model can read and understand text in Devanagari script directly from images, eliminating the need for separate OCR.
    \item \textbf{Strong multilingual capabilities}: Pre-training on diverse multilingual data enables transfer to low-resource languages like Nepali.
    \item \textbf{Dynamic resolution}: The model handles images of varying sizes (min: $256 \times 28^2$, max: $512 \times 28^2$ pixels) without fixed resizing.
    \item \textbf{Instruction following}: The model can follow complex task instructions, enabling prompt-based classification.
\end{itemize}

\subsection{Stage 1: Generative Fine-Tuning with LoRA}

In Stage 1, we fine-tune the model to generate class labels directly from meme images. This generative approach leverages the model's pre-trained language understanding capabilities.

\subsubsection{Low-Rank Adaptation (LoRA)}

Rather than fine-tuning all parameters, we apply LoRA \citep{hu2022lora} to inject trainable low-rank matrices into the model. For a pre-trained weight matrix $W_0 \in \mathbb{R}^{d \times k}$, LoRA adds:
\begin{equation}
W = W_0 + \Delta W = W_0 + BA
\end{equation}
where $B \in \mathbb{R}^{d \times r}$, $A \in \mathbb{R}^{r \times k}$, and $r \ll \min(d, k)$ is the rank. The scaling factor $\alpha / r$ modulates the contribution of the adaptation.
Table~\ref{tab:lora} shows task-specific LoRA configurations.

\begin{table}[htbp]
\centering
\small
\begin{tabular}{lcc}
\toprule
\textbf{Parameter} & \textbf{Hate Speech} & \textbf{Sentiment} \\
\midrule
Rank ($r$) & 64 & 32 \\
Alpha ($\alpha$) & 128 & 64 \\
Scaling ($\alpha/r$) & 2.0 & 2.0 \\
Dropout & 0.05 & 0.1 \\
Trainable params & $\sim$0.5\% & $\sim$0.25\% \\
\bottomrule
\end{tabular}
\caption{LoRA configurations. Sentiment uses smaller rank and higher dropout to prevent overfitting on the three-class problem.}
\label{tab:lora}
\end{table}

\subsubsection{Task-Specific Prompts}

We design prompts that establish the model's role and provide clear classification criteria.
\vspace{1em}

\noindent\textbf{Hate Speech Detection Prompts:}

\vspace{0.8em}
\noindent\textbf{System prompt:} \textit{``You are an expert at detecting hate speech in Nepali memes. Analyze both the visual content and any Nepali/Devanagari text in the image. Consider cultural context, targeted groups (ethnicity, religion, gender, caste), and implicit meanings.''}

\vspace{0.8em}
\noindent\textbf{User prompt:} \textit{``Is this Nepali meme hateful?''}

\vspace{0.4em}
\noindent\textbf{Categories:}
\begin{itemize}
    \setlength{\itemsep}{4pt}
    \setlength{\topsep}{4pt}
    \item \textit{Hate: Content that targets or attacks groups based on ethnicity, religion, gender, caste, or promotes violence/discrimination}
    \item \textit{Safe: Non-hateful content (neutral, positive, jokes without targeting groups)}
\end{itemize}

\vspace{0.4em}
\noindent\textit{Answer with one word only: Hate or Safe.}

\noindent\textit{Answer:}
\vspace{1em}

\noindent\textbf{Sentiment Analysis Prompts:}

\vspace{0.8em}
\noindent\textbf{System prompt:} \textit{``You are an expert at analyzing sentiment in Nepali memes and social media content. You understand Nepali culture, humor, sarcasm, and emotional expression. Analyze both the visual elements (images, facial expressions, symbols) and Devanagari/English text. Pay attention to cultural references, meme templates, and contextual meaning.''}

\vspace{0.8em}
\noindent\textbf{User prompt:} \textit{``Analyze the overall sentiment of this Nepali meme.''}

\vspace{0.4em}
\noindent\textbf{Categories:}
\begin{itemize}
    \setlength{\itemsep}{4pt}
    \setlength{\topsep}{4pt}
    \item \textit{Negative: Expresses criticism, mockery, sadness, anger, frustration, disappointment, or pessimism}
    \item \textit{Neutral: Factual information, observations, or content without clear emotional valence}
    \item \textit{Positive: Expresses joy, humor, encouragement, celebration, love, hope, or optimism}
\end{itemize}

\vspace{0.4em}
\noindent\textit{Consider: (1) Visual content and facial expressions, (2) Text meaning (literal and contextual), (3) Cultural context and references, (4) Overall emotional tone.}

\vspace{0.6em}
\noindent\textit{Respond with ONLY one word: Negative, Neutral, or Positive.}

\vspace{0.4em}
\noindent\textit{Answer:}
\vspace{1em}
\subsubsection{MLP Projection Head}

Alongside LoRA, we train an MLP projection head that maps the model's last hidden state to a lower-dimensional embedding space:
\begin{align}
\mathbf{h}_1 &= \text{GELU}(W_1 \mathbf{h} + b_1) \\
\mathbf{h}_2 &= \text{LayerNorm}(\mathbf{h}_1) \\
\mathbf{z} &= W_2 \mathbf{h}_2 + b_2
\end{align}
where $\mathbf{h} \in \mathbb{R}^{4096}$ is the last hidden state, $W_1 \in \mathbb{R}^{2048 \times 4096}$, $W_2 \in \mathbb{R}^{512 \times 2048}$, and $\mathbf{z} \in \mathbb{R}^{512}$ is the projected embedding.

This projection head serves two purposes: (1) dimensionality reduction for efficient contrastive learning in Stage 2, and (2) learning task-specific representations beyond what the language model head captures.

\subsubsection{Training Objective}

The Stage 1 training objective is the standard language modeling loss, but computed only over the target tokens. Given a prompt $x$ and target label $y$ (e.g., ``Hate'' or ``Safe''), we minimize:
\begin{equation}
\mathcal{L}_{\text{LM}} = -\log P(y | x, \text{image})
\end{equation}

We mask the prompt tokens (set labels to -100) so that gradients flow only through the answer tokens. This focuses learning on the classification decision.

\subsubsection{Class Imbalance Handling}

Both datasets exhibit significant class imbalance. We employ three complementary strategies:

\paragraph{Minority Class Oversampling:} We duplicate minority class samples to achieve balanced class frequencies. For hate speech, we target a 1:1 ratio (hate:safe). For sentiment, we target 1:1:1 (negative:neutral:positive). The majority class is not downsampled to preserve all available information.

\paragraph{Image Augmentation:} Oversampled (duplicated) images undergo random transformations to increase diversity and prevent the model from memorizing specific samples. With 80\% probability, we apply:
\begin{itemize}
    \item Horizontal flip: 50\% probability
    \item Rotation: uniform in $[-15\degree, +15\degree]$
    \item Brightness adjustment: factor in $[0.8, 1.2]$
    \item Contrast adjustment: factor in $[0.8, 1.2]$
    \item Saturation adjustment: factor in $[0.8, 1.2]$
\end{itemize}
Each transformation is applied independently with 50\% probability, creating diverse augmented versions.

\paragraph{Focal Loss (Sentiment Only):} For the three-class sentiment task, we additionally use focal loss \citep{lin2017focal}:
\begin{equation}
\mathcal{L}_{\text{focal}} = -\alpha_t (1 - p_t)^\gamma \log(p_t)
\end{equation}
with focusing parameter $\gamma = 2.0$. The modulating factor $(1 - p_t)^\gamma$ reduces the loss contribution from easy examples, focusing training on hard cases. Class weights $\alpha_t$ are set to inverse class frequencies.

\subsubsection{Training Configuration}

\begin{table}[htbp]
\centering
\small
\setlength{\tabcolsep}{4pt}
\begin{tabular}{p{2.2cm}cc}
\toprule
\textbf{Parameter} & \textbf{Hate Speech} & \textbf{Sentiment} \\
\midrule
Learning rate & $5 \times 10^{-5}$ & $1 \times 10^{-4}$ \\
Batch size & 2 & 2 \\
Gradient accum. & 8 & 12 \\
Effective batch & 16 & 24 \\
Epochs & 5 & 5 \\
Optimizer & \multicolumn{2}{c}{AdamW ($\beta_1$=0.9, $\beta_2$=0.999)} \\
Weight decay & \multicolumn{2}{c}{0.01} \\
Epsilon & \multicolumn{2}{c}{$10^{-8}$} \\
LR scheduler & \multicolumn{2}{c}{Cosine, 10\% warmup} \\
Grad. clipping & \multicolumn{2}{c}{1.0} \\
Precision & \multicolumn{2}{c}{bfloat16 mixed} \\
\bottomrule
\end{tabular}
\caption{Stage 1 training configurations.}
\label{tab:stage1config}
\end{table}

The sentiment task uses a higher learning rate and larger effective batch size. We found this necessary for the more complex three-class decision boundary.

\subsection{Stage 2: Contrastive Backbone Fine-Tuning}

Stage 2 continues training the LoRA layers and MLP with a joint classification and contrastive objective. The goal is to learn embeddings where same-class samples cluster together while different-class samples are pushed apart.

\subsubsection{Linear Classifier}

We add a linear classifier on the projected embeddings:
\begin{equation}
\mathbf{y} = W_c \mathbf{z} + b_c
\end{equation}
where $W_c \in \mathbb{R}^{C \times 512}$, $b_c \in \mathbb{R}^{C}$, and $C$ is the number of classes (2 for hate speech, 3 for sentiment).

\subsubsection{Supervised Contrastive Loss}

We use a supervised variant of InfoNCE loss \citep{oord2018representation, khosla2020supervised}. For each anchor embedding $\mathbf{z}_i$ with label $y_i$, we sample $k^+$ positive embeddings (same label) and $k^-$ negative embeddings (different labels) from the training set.

The contrastive loss is:
\begin{equation}
\mathcal{L}_{\text{con}} = -\frac{1}{k^+} \sum_{j=1}^{k^+} \log \frac{\exp(s_{ij}^+ / \tau)}{D_i}
\end{equation}
where $s_{ij}^+ = \text{sim}(\mathbf{z}_i, \mathbf{z}_j^+)$ is the cosine similarity between anchor $i$ and positive $j$, and the denominator is:
\begin{equation}
D_i = \sum_{p=1}^{k^+} \exp(s_{ip}^+ / \tau) + \sum_{n=1}^{k^-} \exp(s_{in}^- / \tau)
\end{equation}

where $\text{sim}(\mathbf{a}, \mathbf{b}) = \mathbf{a}^\top \mathbf{b} / (\|\mathbf{a}\| \|\mathbf{b}\|)$ is cosine similarity and $\tau = 0.07$ is the temperature parameter. All embeddings are L2-normalized before computing similarity.

The temperature $\tau$ controls the concentration of the distribution. Lower temperatures make the model more sensitive to hard negatives but can lead to training instability. We use $\tau = 0.07$ following prior work \citep{khosla2020supervised}.

\subsubsection{Joint Training Objective}

The Stage 2 loss combines classification and contrastive terms:
\begin{equation}
\mathcal{L} = \lambda_{\text{cls}} \mathcal{L}_{\text{cls}} + \lambda_{\text{con}} \mathcal{L}_{\text{con}}
\end{equation}

For hate speech: $\lambda_{\text{cls}} = 1.0$, $\lambda_{\text{con}} = 0.3$. The classification loss uses label smoothing of 0.05 to prevent overconfident predictions.

For sentiment: $\lambda_{\text{cls}} = 1.0$, $\lambda_{\text{con}} = 0.4$. The classification loss uses inverse-frequency class weights to handle imbalance.

\subsubsection{Stage 2 Training Configuration}

Table~\ref{tab:stage2config} shows Stage 2 hyperparameters. A key difference from Stage 1 is the much lower backbone learning rate and stronger regularization, especially for sentiment.
\begin{table}[htbp]
\centering
\small
\setlength{\tabcolsep}{4pt}
\begin{tabular}{p{2.4cm}cc}
\toprule
\textbf{Parameter} & \textbf{Hate Speech} & \textbf{Sentiment} \\
\midrule
Backbone LR & $2 \times 10^{-5}$ & $5 \times 10^{-6}$ \\
Classifier LR & $1 \times 10^{-4}$ & $2 \times 10^{-5}$ \\
Batch size & 2 & 2 \\
Gradient accum. & 8 & 12 \\
Epochs & 5 & 3 \\
Weight decay & 0.01 & 0.1 \\
Early stopping & -- & 2 \\
Positives ($k^+$) & 3 & 4 \\
Negatives ($k^-$) & 5 & 6 \\
Temperature ($\tau$) & 0.07 & 0.07 \\
$\lambda_{\text{con}}$ & 0.3 & 0.4 \\
Label smoothing & 0.05 & -- \\
Class weights & -- & Inverse freq. \\
\bottomrule
\end{tabular}
\caption{Stage 2 training configurations. Sentiment uses aggressive regularization to prevent overfitting.}
\label{tab:stage2config}
\end{table}

The sentiment task required substantially different settings:
\begin{itemize}
    \item 4$\times$ lower backbone learning rate ($5 \times 10^{-6}$ vs $2 \times 10^{-5}$)
    \item 10$\times$ higher weight decay (0.1 vs 0.01)
    \item Fewer epochs (3 vs 5) with early stopping
\end{itemize}

These aggressive regularization choices were necessary because the three-class sentiment task has a smaller effective training set per class and showed signs of overfitting early in our experiments.

\subsection{Inference and Ensemble}

At inference time, we combine predictions from both stages to leverage their complementary strengths.

\subsubsection{Stage 1 Predictions}

We extract token-level probabilities from the language model head. For hate speech:
\begin{equation}
p_{\text{hate}}^{(1)} = \frac{\exp(l_{\text{Hate}})}{\exp(l_{\text{Hate}}) + \exp(l_{\text{Safe}})}
\end{equation}
where $l_{\text{Hate}}$ and $l_{\text{Safe}}$ are logits for the respective tokens at the first generated position.

For sentiment, we similarly compute:
\begin{equation}
p_c^{(1)} = \frac{\exp(l_c)}{\sum_{c'} \exp(l_{c'})}
\end{equation}
for each class $c \in \{\text{Negative}, \text{Neutral}, \text{Positive}\}$.

\subsubsection{Stage 2 Predictions}

We apply softmax to the linear classifier outputs:
\begin{equation}
p^{(2)} = \text{softmax}(W_c \mathbf{z} + b_c)
\end{equation}

\subsubsection{Ensemble Strategy}

The final prediction is a weighted combination:
\begin{equation}
p_{\text{ensemble}} = w \cdot p^{(1)} + (1 - w) \cdot p^{(2)}
\end{equation}

We tune the ensemble weight $w$ and decision threshold jointly via grid search on the validation set, optimizing for macro F1. For hate speech:
\begin{itemize}
    \item Ensemble weight $w$: search in $[0.1, 0.9]$ with step 0.02
    \item Decision threshold: search in $[0.25, 0.75]$ with step 0.005
\end{itemize}

For sentiment (multi-class), we use argmax on the ensembled probabilities and only tune the ensemble weight.

\section{Experimental Setup}

\subsection{Implementation Details}

We implement our system using:
\begin{itemize}
    \item PyTorch 2.4.0 with CUDA
    \item HuggingFace Transformers $\geq$ 4.46.0 \citep{wolf2020transformers}
    \item PEFT $\geq$ 0.13.0 for LoRA \citep{peft2024}
    \item Accelerate $\geq$ 1.0.0 for distributed training
\end{itemize}

Training runs on NVIDIA H100 GPUs via Modal Labs cloud infrastructure. Each training run takes approximately 2--4 hours per stage.

\subsection{Reproducibility}

We use fixed random seeds (42) for data splitting, model initialization, and training. All hyperparameters are specified in Tables~\ref{tab:lora}--\ref{tab:stage2config}. Our code will be made available upon publication.

\section{Results and Discussion}

\subsection{Official Results}

\begin{table}[htbp]
\centering
\begin{tabular}{lcc}
\toprule
\textbf{Subtask} & \textbf{Macro F1} & \textbf{Rank} \\
\midrule
A: Hate Speech Detection & 0.7970 & 2nd \\
B: Sentiment Analysis & 0.5177 & 4th \\
\bottomrule
\end{tabular}
\caption{Official leaderboard rankings on CHiPSAL 2026.}
\label{tab:results}
\end{table}

Table~\ref{tab:results} shows our official results. We achieved strong performance on hate speech detection with F1 of 0.797, securing 2nd place. For sentiment analysis, we achieved F1 of 0.518, placing 4th. The binary hate speech task proved more tractable, likely due to the clearer class boundary compared to the three-way sentiment distinction.

\subsection{Ablation Study}
\begin{table}[htbp]
\centering
\small
\begin{tabular}{lc}
\toprule
\textbf{Configuration} & \textbf{Test F1} \\
\midrule
\multicolumn{2}{l}{\textit{Subtask A: Hate Speech Detection}} \\
Stage 1 (LoRA baseline) & 0.74 \\
\quad + Oversampling & 0.76 \\
\quad + Image augmentation & 0.77 \\
\quad + Stage 2 contrastive & 0.79 \\
\quad + Threshold tuning & \textbf{0.80} \\
\midrule
\multicolumn{2}{l}{\textit{Subtask B: Sentiment Analysis}} \\
Stage 1 (LoRA baseline) & 0.49 \\
\quad + Focal loss & 0.50 \\
\quad + Oversampling \& augmentation & 0.51 \\
\quad + Stage 2 contrastive & \textbf{0.52} \\
\bottomrule
\end{tabular}
\caption{Ablation study. Each row adds one component cumulatively. F1 scores are reported on the held-out test set, except threshold tuning which was optimized on the validation split and then applied to the test set.}
\label{tab:ablation}
\end{table}

Table~\ref{tab:ablation} presents detailed ablation results. Key findings:

\paragraph{LoRA scope matters.} Applying LoRA to all attention and feed-forward layers outperformed applying it only to attention layers (+3 points on hate speech).

\paragraph{Class imbalance handling is crucial.} The combination of oversampling and augmentation contributes 2--3 points. Focal loss adds another point for the three-class sentiment task by addressing neutral class bias.

\paragraph{Contrastive learning provides consistent gains.} Stage 2 adds 2 points on hate speech and contributes to sentiment, demonstrating that the contrastive objective learns more discriminative representations.

\paragraph{Threshold tuning improves hate speech detection.} Optimizing the classification threshold on the validation split and applying it to the test set adds 1 point, reaching the final score of 0.80.

\subsection{Qualitative Examples}

\paragraph{Successful Cases:} The model correctly identifies explicit hate speech with visual symbols (e.g., derogatory depictions) combined with text. For sentiment, clear emotional expressions in both image and text (e.g., smiling faces with celebratory text) are reliably classified.

\paragraph{Failure Cases:} A meme showing a historical figure with a quote that is hateful in Nepali cultural context but neutral in isolation was misclassified as safe. The model lacks sufficient cultural knowledge to interpret such references.

\section{Conclusion}

We presented a two-stage vision-language system for Nepali meme classification at CHiPSAL 2026. By combining LoRA fine-tuning with contrastive learning on Qwen3-VL-8B-Instruct, we achieved 2nd place on hate speech detection (F1: 0.797) and 4th place on sentiment analysis (F1: 0.518).

Our results highlight several key findings. Modern VLMs with native Devanagari support enable end-to-end processing of South Asian language memes without OCR pipelines, eliminating error propagation from intermediate steps. Two-stage training (generative then contrastive) outperforms either approach alone by 3--5 F1 points, demonstrating the complementarity of language modeling and metric learning objectives. Task-specific hyperparameter tuning proved essential, as sentiment required much stronger regularization than hate speech. Finally, careful handling of class imbalance is crucial for macro F1 optimization, contributing up to 7 points across our ablation.

Promising future directions include retrieval-augmented approaches to incorporate cultural knowledge, synthetic data augmentation to expand limited training sets, multi-task learning jointly across hate speech and sentiment, and explainability analysis to better understand model decisions for culturally sensitive content.

\section{Limitations}

\begin{itemize}
    \item \textbf{Data scarcity:} With approximately 1,000 samples per task, overfitting remains a concern. Our validation results may not fully generalize to the test distribution. Due to the high computational cost of each training run, we did not evaluate variance across multiple random seeds; results may vary with different initializations.
    \item \textbf{Cultural context:} Nepali cultural references, historical context, and humor styles may not be well-represented in Qwen3-VL's pretraining data, limiting performance on culturally-specific content.
    \item \textbf{Prompt sensitivity:} We did not exhaustively search prompt variations, which could further improve performance.
    \item \textbf{Code-mixing:} While Qwen3-VL handles Nepali-English code-mixing to some extent, highly mixed content may still pose challenges.
    \item \textbf{Baseline comparisons:} We did not evaluate Qwen3-VL-8B-Instruct in a zero-shot (no fine-tuning) setting or compare against an OCR + text-only pipeline, which would better quantify the contribution of our end-to-end multimodal approach. We leave these comparisons for future work.
\end{itemize}

\section{Ethical Considerations}

Hate speech detection systems must be deployed carefully to avoid over-censorship or bias. Our model may have false positives on legitimate political speech or cultural content unfamiliar to the base model. We recommend human review for content moderation decisions.

\section{Acknowledgements}

We thank the CHiPSAL 2026 organizers for creating this valuable benchmark and shared task infrastructure. We acknowledge Modal Labs for computational resources.

\section{Bibliographical References}\label{sec:reference}

\bibliographystyle{lrec2026-natbib}
\bibliography{chipsal}

\end{document}